\documentclass[cameraready]{Interspeech}

\title{MoDiCoL: A Modular Diagnostic Continual Learning Dataset for Robust Speech Recognition}

\author[orcid=0000-0002-2847-5097]{Theresa}{Pekarek Rosin}
\author[orcid=0000-0002-1378-0435]{Matthias}{Kerzel}
\author[orcid=0000-0003-1343-4775]{Stefan}{Wermter}

\address{
    Knowledge Technology, Department of Informatics, University of Hamburg, 
    Germany
}

\email{\{theresa.pekarek-rosin, matthias.kerzel, stefan.wermter\}@uni-hamburg.de}

\keywords{speech recognition, continual learning, audio dataset, speech resources}

\usepackage{comment}

\begin{document}

\maketitle

\begin{abstract}
    Modern Automatic Speech Recognition (ASR) systems have made remarkable progress on standard benchmarks, yet performance gaps have emerged under real-world distribution shifts, caused by recording conditions, accents, speech impairments, and noise. Existing datasets and benchmarks typically isolate these factors, which overlooks their co-occurrence in real-world applications. In this paper, we argue that model robustness can be treated as a dynamic capability that continually develops, and we introduce MoDiCoL, a Modular Diagnostic Continual Learning dataset designed for controlled analysis of linguistic content, speaker characteristics, and acoustic environments. Furthermore, we propose a real-world-inspired continual learning curriculum to simulate incremental updates and study how robustness is acquired, transferred, and forgotten. We evaluate three continual learning strategies and provide detailed insights into robustness under evolving conditions.
\end{abstract}

\section{Introduction}
Modern Automatic Speech Recognition (ASR) models have improved significantly on standard benchmarks over the last few years~\cite{Radford2022}, yet they still struggle with speech that differs from the standard, fixed setups. Challenges include impaired or disfluent speech~\cite{mujtaba-etal-2024-lost}, accents or dialects~\cite{Dhanjal2023}, varying noise or environments~\cite{Vinnikov2024}, and elderly or children~\cite{FENG2024101567} speakers. Even after the utilization of large-scale datasets for pretraining, these performance gaps still persist~\cite{hwang-etal-2025-evaluating}.

Existing benchmarks or datasets are typically built for specific factors, e.g., a model’s robustness to noise~\cite{Vinnikov2024,reddy2019scalable}, accents~\cite{svarah_dataset}, or impairment~\cite{torgo,Cleland2019TheIO}, and evaluated separately.
This does not reflect real-world conditions, where such distribution shifts accumulate over time. Even datasets with rich metadata, built for evaluation, such as Casual Conversations v2~\cite {Porgali_2023_CVPR}, remain limited by uncontrollable factors, such as recording conditions and speaker selection, making it harder to isolate the factor with the greatest impact.
Additionally, since real-world applications often involve accumulating changes, we argue that robustness could be studied as a continual capability that evolves across sequential distribution shifts.

Continual learning (CL) is a sequential learning paradigm that preserves previously acquired knowledge~\cite{PKPKW19,cl_survey}. Therefore, CL enables a simulation of real-world incremental model updates and an examination of how robustness is acquired during training and whether it can be transferred across different factor drifts. While CL has been applied to speech for domain adaptation, such as accent~\cite{vandereeckt23_interspeech} and language adaptation~\cite{dellalibera2023}, its potential as a diagnostic method to identify where forgetting occurs in pretrained ASR models has not yet been explored in the same way as it has been, for example, for Vision-Language models~\cite{ABLW23a}.

To address these identified knowledge gaps, we introduce \textbf{MoDiCoL}\footnote{\url{https://huggingface.co/datasets/TPekarekRosin/modicol}}, a \textbf{Mo}dular \textbf{Di}agnostic \textbf{Co}ntinual \textbf{L}earning dataset, in which we offer a controlled evaluation of different factors that influence ASR performance: linguistic content, speaker characteristics, and acoustic environment. We implement a real-world-inspired CL curriculum for the dataset, using three different CL strategies to analyze the causes and effects of distribution drift across these factors and mitigate the resulting robustness degradation.

\section{MoDiCoL: A modular diagnostic continual learning dataset}
MoDiCoL is a speech dataset designed to study the robustness of ASR models to different drift factors in a controlled, continual setting. We construct MoDiCoL using a systematic factorial design that enables a rigorous evaluation of linguistic, speaker, and acoustic variation with clearly defined experimental runs. By combining real-world and synthetic speech with a configuration-dependent augmentation pipeline, MoDiCoL allows a targeted evaluation of how different types of factorial drift affect model adaptation and forgetting.

\subsection{Factors}
As shown in Table~\ref{tab:domains}, MoDiCoL covers three factor categories: Linguistic Content, Speaker Characteristics, and Acoustic Environment. Each of these categories describes a set of factors expressed at different levels.

\begingroup
    \setlength{\tabcolsep}{6pt} 
\renewcommand{\arraystretch}{1.1} 
\begin{table}[t]
\caption{The factors covered in MoDiCoL and their levels. Linguistic content describes the factors that account for variation in speaking patterns and vocabulary within the recording context. Speaker characteristic factors cover the speaker-dependent variations, and acoustic environment factors describe the recording environment.}
    \centering
    \begin{tabular}{|l|l|}
    \hline
        \multicolumn{2}{|l|}{\textsc{Linguistic Content}} \\ \hline
         Vocabulary Domain & Medical, Air Traffic Control, Other\\ 
         Speech Style & Read, Spontaneous, Conversational \\ \hline
         \multicolumn{2}{|l|}{\textsc{Speaker Characteristics}} \\ \hline
         Age & Child, Adult, Elderly \\
         Accent & English, South-East Asian, Other \\
         Health & Healthy, Impaired \\
         Pauses & Yes, No \\
         Disfluencies & Yes, No \\ \hline
         \multicolumn{2}{|l|}{\textsc{Acoustic Environment}} \\ \hline
         Noise Type & Clean, Babble, Fan \\
         SNR Level & Clean, 10dB, 20dB \\
         Distance & Close, Far \\ \hline
    \end{tabular}
    \label{tab:domains}
\end{table}
\endgroup

\textbf{Linguistic content} refers to the language-related characteristics of the dataset. This includes the vocabulary domain, which specifies the subject area (medical, air traffic control, or other domains), and the speech style, which captures how speech is produced (read, spontaneous, or conversational).

\textbf{Speaker characteristics} describe demographic and speech-production factors of the respective speakers in the dataset. These include age group (child, adult, elderly), accent (English, South-East Asian, or other), and health (healthy or impaired). Additionally, we include pauses and disfluencies as separate factors, since they also occur in healthy speech.

\textbf{Acoustic environment} describes the recording conditions under which the speech data was captured. It includes the noise type (clean, babble, or fan noise), the signal-to-noise ratio (SNR) level (clean, 10dB, or 20dB), and the distance between the speaker and the microphone.

\subsection{Structure}
We create the dataset with an L27 orthogonal array (OA) following the Taguchi design method~\cite{taguchi1993} and two foldover dimensions $F_0$ and $F_1$ to accommodate the six three-dimensional factors (vocabulary domain, speech style, age, accent, noise type, and SNR level) and the four two-dimensional factors (health, pauses, disfluencies, and distance).

The foldover dimensions create four slices, which we use to perform cyclic permutation, yielding 108 ($27\times4$) distinct run configurations in total.
We fill each of these runs with 75 samples, a standard heuristic for mixed-level fractional factorial designs~\cite{montgomery2008design}, resulting in a dataset of \textbf{8100} samples overall. Within these runs, some factor-level combinations require special handling:
\begin{enumerate}
    \item medical (vocabulary domain), elderly/child (age)
    \item ATC (vocabulary domain), elderly/child (age)
    \item medical/ATC (vocabulary domain), impaired (health)
    \item clean (SNR level) and babble/fan (noise type)
\end{enumerate}

We differentiate between combinations unlikely to occur in real-world datasets, but that can still provide insight into model behavior (1–3) and truly infeasible combinations that violate logical constraints (4). The first type is accommodated using synthetic speech generation and voice cloning, while we repair the experimental design by changing the noise type to ``clean" for the second type.

\subsection{Dataset compilation}\label{sec:datasetcomp}
The OA of the dataset contains many combinations that lack available real-world data (e.g., impaired speech with non-English accents). We therefore populate as many runs as possible with open-source speech datasets and use text datasets to generate synthetic speech for missing combinations.

Healthy children’s speech is sourced from Non-native Children English Speech (NNCES)~\cite{nnces}, Children Speech Recordings~\cite{children_english}, and Ultrax Typically Developing Children (UXTD)~\cite{eshky18_interspeech,ribeiro19_interspeech}, while impaired children’s speech comes from the Cleft dataset~\cite{Cleland2019TheIO}.
Healthy adult and elderly speech is taken from Svarah~\cite{svarah_dataset}, Trustworthy Intent in Speech (TIS)~\cite{trustworthy_dataset}, TED-LIUMv2~\cite{tedlium2}, and Common Voice~\cite{Ardila2019}, with DailyDialog~\cite{dailydialog} providing conversational data.

For domain-specific speech, ATC data is obtained from the ATCO-2 1-hour test split~\cite{atco2} and the UWB-ATCC dataset~\cite{uwbatc}, which we also use as transcripts for the synthetic speech generation alongside related texts from Wikipedia.
Medical speech is collected from United-Syn-Med~\cite{united_medsyn} and the Eka Medical ASR Evaluation Dataset~\cite{eka}, supplemented with medical texts from MedRAG~\cite{medrag}, Medical Transcriptions~\cite{medtranscripts}, MedDialog~\cite{zeng-etal-2020-meddialog}, NoteChat~\cite{wang-etal-2024-notechat}, and simulated medical exams~\cite{Fareez2022-ts}.

The synthetic speech is generated using XTTS-v2~\cite{casanova24_interspeech}, a multilingual zero-shot text-to-speech (TTS) model that uses reference audio to match speaker characteristics. We collect such reference files across age, accent, $f_0$, and health conditions from datasets including the Perceptual Voice Quality Database~\cite{pvqd}, the Voice Cloning Toolkit (VCTK)~\cite{vctk}, Cleft~\cite{Cleland2019TheIO}, Common Voice 11.0~\cite{Ardila2019}, Children Speech Recordings~\cite{children_english}, NNCES~\cite{nnces}, Ultrax Speech Sound Disorders (UXSSD)~\cite{eshky18_interspeech}, and TORGO~\cite{torgo}.

Overall, MoDiCoL contains \SI{18.79}{\hour} of speech, of which \SI{14.08}{\hour} are synthetic, with an average sample length of \SI{8.35}{\second}.

\subsection{Augmentation pipeline}

We resample all files to \SI{16}{\kilo\hertz} and save them as wave files.
To control for background noise and other aspects of the acoustic environment, as well as disfluencies and pauses, we augment both real and generated speech samples using the following pipeline, as specified in the respective run configuration.
\begin{enumerate}
    \item \textbf{Denoising.} We first suppress background noise for all samples using a DNN-HMM hybrid system~\cite{bagchi2018spectral}, initialized from a publicly available checkpoint on Hugging Face\footnote{\url{https://huggingface.co/speechbrain/mtl-mimic-voicebank}}.
    \item \textbf{Disfluency Insertion.} If no natural disfluencies are present, we synthesize filler words using the XTTSv2 model, conditioning on the original audio to preserve speaker identity. Filler words are inserted using the same procedure as pause insertion (see Step 4).
    \item \textbf{Impairment Simulation.} For speakers with South-East Asian accents and impaired speech, suitable open-source reference recordings were unavailable. In these cases, we either use the healthy reference voices for TTS or start from healthy real speech and simulate impairment. Impairment is introduced via prosodic and spectral modifications, including tremor, jitter, shimmer, and temporal perturbations.
    \item \textbf{Pause Insertion/Removal.} If the file contains no pauses, they are inserted by detecting existing silent segments (e.g., between words) or, if none are available, by randomly selecting insertion points. If pauses are disabled in the configuration, noticeable silent segments are removed instead.
    \item \textbf{Distance Simulation.} To simulate increased distance between speaker and microphone, we add reverberation, effectively modeling a larger room and a longer acoustic path.
    \item \textbf{Noise Injection.} Background noise is added according to the predefined signal-to-noise ratio (SNR) and noise type. For babble and fan noise, we sample from the Microsoft Scalable Noisy Speech Dataset (MS-SNSD)~\cite{reddy2019scalable}, which provides dedicated noise recordings for each type. For clean conditions, synthetic random noise is generated. The selected noise is mixed with the speech signal at the specified SNR level.
\end{enumerate}

\section{Methodology}

\subsection{Continual learning curriculum}
Continual learning (CL) can be defined as follows:
$\mathcal{T}$ is the set of tasks that is learned by the model in sequence $\mathcal{T}:={t_1, t_2, …, t_N}$, where $\mathcal{D}={\mathcal{D}_1, \mathcal{D}_2, …, \mathcal{D}_N}$ is the corresponding stream of data with unknown distributions over an input-output space $\mathcal{X}\times \mathcal{Y}$ \cite{cl_survey,vandeven2024clcf}. The model receives a batch of samples at each step, belonging to the current task $t_i$, from data $\mathcal{D}_t$ with its distinct distribution.

We sample from LibriSpeech~\cite{librispeech} to define a control setting $t_0$ against which all drift types are compared: adult, healthy, read, no pauses, no disfluencies, clean, close, other vocabulary, English accent. We then define a CL curriculum of disjoint run sets that represent different types of data drift:

\begin{itemize}
    \item $t_1$: \textbf{Acoustic Environment Drift}. Introduction of environmental variation through babble and fan noises, varied SNR levels, and microphone distances.
    \item $t_2$: \textbf{Speaker Characteristics Drift}. Inclusion of new speaker groups, i.e., children, the elderly, and impaired speakers with different accents. Acoustic conditions are kept moderate.
    \item $t_3$: \textbf{Linguistic Content Drift}. Exposure to new domains and speaking styles, i.e., spontaneous and conversational speech, and specialized vocabulary (medical, air traffic control).
    \item $t_4$: \textbf{Compound Drift}. Combination of multiple drift types: elderly or impaired speakers at a far distance in noisy environments, with conversational or domain-specific speech.
\end{itemize}

To study robustness transfer across drift types, we evaluate whether robustness acquired on one drift generalizes to others and whether curriculum order influences model performance. We generate two additional permutations of the displayed curriculum for the evaluation while keeping task $t_4$ fixed.

\subsection{Experimental setup}
In our experimental setup, we employ three continual learning (CL) methods to analyze different aspects of sequential drift: rehearsal-based stabilization with Experience Replay~\cite{experiencereplay}, representation preservation with Representation-level Regularization~\cite{POMPONI2020139}, and gradient subspace isolation with Orthogonal Gradient Descent~\cite{ogd}. As the backbone model, we use a pretrained \texttt{whisper-small.en} model~\cite{Radford2022}.

\textbf{Experience Replay (ER)}~\cite{experiencereplay} is a rehearsal-based approach that maintains a memory buffer containing a fraction of data from previous tasks. During fine-tuning on a new task, buffered samples are interleaved with the current training data. We evaluate small (5\%) and medium (10\%) buffer sizes, with an equal number of samples per past task, to assess whether stability regarding drifts depends on rehearsal mechanisms.

For \textbf{Representation-level Regularization (RLR)}~\cite{POMPONI2020139}, we introduce a regularization term that constrains the encoder to preserve previously learned representations. We retain a frozen copy of the encoder after each task and minimize the cosine similarity loss between the mean-pooled encoder outputs of the current and the frozen models. Using RLR, we assess how the curriculum affects the model’s acoustic representation and whether forgetting arises from representational drift.

\textbf{Orthogonal Gradient Descent (OGD)}~\cite{ogd} mitigates forgetting by constraining parameter updates to directions that do not interfere with prior tasks. We apply OGD to the encoder and compute an average gradient over a small subset of the task data, then normalize it to obtain the task-specific direction. During training, gradients are projected onto the orthogonal complement of all stored directions to prevent interference through parameter updates. We measure task similarity by computing the cosine similarity between current gradients and stored directions to quantify geometric overlap between gradient subspaces and entanglement between drift factors.

Performance across tasks is measured using Average Word Error Rate (A-WER) and Average Incremental WER (AI-WER). Memory stability is assessed via Forgetting Measure (FM) and Backward Transfer (BWT), while learning plasticity is quantified using Forward Transfer (FWT) and Intransigence Measure (IM), following Wang et al.~\cite{cl_survey}. Since we use an error rate as the evaluation metric, the interpretation of some metrics changes, which we indicate accordingly.

To approximate real-world conditions for model adaptation, we perform these evaluations in an \textbf{online} and \textbf{streaming} CL setting, where the batch size is limited to one, and the model sees every sample only once~\cite{vandeven2024clcf}. We split each drift into train (70\%), validation (20\%), and test (10\%) sets and train with a learning rate of $1e-5$.
And to quantify the impact of distribution drift, we include a sequential \textbf{fine-tuning} (FT) lower baseline without any CL mechanism, which captures task interference, and a \textbf{joint training} upper baseline, in which all drift splits are trained simultaneously.

\begingroup
    \setlength{\tabcolsep}{6pt} 
\renewcommand{\arraystretch}{1.1} 
\begin{table}[t]
\caption{The WER and BERTScore (F1) results of evaluating MoDiCoL on \texttt{whisper-small.en} \textbf{before} the curriculum training. 
}
    \centering
    \begin{tabular}{lcccc}
    \hline
                    & \multicolumn{2}{c}{\textsc{A-WER} ($\downarrow$)} & \multicolumn{2}{c}{\textsc{F1} ($\uparrow$)}\\
        \textsc{Set}          & \textsc{mean} & \textsc{median} & \textsc{mean} & \textsc{median} \\ \hline
         $t_0$      &  7.42 & 0.0 & 98.28 & 99.90 \\
         $t_1$      & 47.62 & 14.29 & 95.80 & 97.03 \\
         $t_2$      & 87.28& 28.57& 92.42& 94.38\\
         $t_3$      & 141.73& 42.86& 89.54& 91.35\\
         $t_4$      & 43.37& 20.00& 94.28& 95.63\\ \hline
         \textsc{Real}       & 86.32 & 25.00 & 92.43 & 94.55\\
         \textsc{Synthetic}  & 69.97 & 13.33 & 95.02 & 96.90\\ \hline
         \textsc{All}        & 75.04 & 16.67 & 94.22 & 96.37 \\ \hline
    \end{tabular}
    \label{tab:quality}
\end{table}
\endgroup
\begingroup
    \setlength{\tabcolsep}{6pt} 
\renewcommand{\arraystretch}{1.1} 
\begin{table*}[t]
\caption{The results of the CL curriculum, averaged over the three permutations ($\pm$ standard deviation). Since WER is used as a performance measure, some metric interpretations are reversed, e.g., positive BWT scores indicate forgetting. We mark the optimal direction next to the name. \textsc{Joint} denotes the joint training baseline, and \textsc{FT} the sequential fine-tuning baseline.   }
    \centering
    \begin{tabular}{lcccccc}
    \hline
        & \textsc{A-WER} ($\downarrow$) & \textsc{AI-WER} ($\downarrow$) & \textsc{FM} (target $=0$) & \textsc{BWT} ($\downarrow$) & \textsc{FWT} ($\downarrow$) & \textsc{IM} ($\uparrow$) \\ \hline
        \textsc{ER-5\%} & $25.75\pm11.32$& $23.40\pm1.42$& $-12.89\pm14.99$& $12.89\pm14.99$&  $-19.51\pm9.04$&  $26.52\pm0.10$\\ 
        \textsc{ER-10\%} & $\mathbf{17.31\pm0.48}$& $22.83\pm3.84$& $\mathbf{-1.95\pm1.20}$& $\mathbf{1.78\pm0.93}$& $\mathbf{-31.31\pm0.51}$&  $26.64\pm0.22$\\ 
        \textsc{RLR} & $34.28\pm8.80$ & $24.30\pm5.31$ & $-22.34\pm13.32$& $22.33\pm13.32$& $-20.95\pm8.78$&  $25.09\pm1.82$\\  
        \textsc{OGD}  & $26.87\pm9.84$ & $\mathbf{21.19\pm1.18}$ & $-12.75\pm14.30$& $12.16\pm14.80$& $-23.81\pm10.25$& $24.87\pm1.26$\\ \hline
        \textsc{Joint}  & $27.24\pm2.25$ & - & - & - & - & - \\ 
       \textsc{FT} & $34.14\pm8.56$& $23.73\pm3.35$& $-24.55\pm10.99$& $24.55\pm10.99$& $-27.85\pm4.25$& $\mathbf{26.88\pm0.33}$\\ \hline
    \end{tabular}
    \label{tab:clcurr}
\end{table*}
\endgroup

\section{Results}
Before the CL curriculum, we first evaluate the dataset’s overall quality by computing WER and F1 scores with BERTScore~\cite{bertscore} as a semantic similarity metric for \texttt{whisper-small.en}. All metrics are reported on the normalized transcripts.

As Table~\ref{tab:quality} shows, the unadapted model performs best on the control setting $t_0$, where it achieves low WER and a high F1 score, but seems to deteriorate substantially under the distribution shifts. For acoustic drift ($t_1$), the performance drops moderately, but degrades even more severely for speaker ($t_2$) and linguistic drift ($t_3$). Here, the WER increases dramatically, indicating a strong sensitivity to changes in speaker characteristics and language content. Interestingly, compound drift ($t_4$), despite combining all previous drift factors, does not produce the worst performance, suggesting that drift effects do not accumulate in difficulty when applied together.

Overall, there appears to be a large difference between the mean-averaged WER and the median WER of the unadapted model. This indicates that the model generally transcribes the audio with a reasonable error rate (16.67\%), but that a few severe outliers are distorting the mean. By examining the transcripts, we see that hallucinations are a major contributor to the high WER, but since the semantic similarity scores (F1) remain high throughout, they do not seem to affect the semantic alignment between the model output and the ground truth. Additionally, synthetic speech generally performs better than real-world speech, likely due to the greater variability in the real-world subset (speakers, impairments, etc.).

The results of the CL curriculum (Table~\ref{tab:clcurr}) show a clear benefit of utilizing continual learning for the introduction of different drift types.
ER with a replay buffer size of 10\% is the most balanced approach, with an A-WER of $17.31\pm0.48$, which outperforms not only the other CL methods but also the joint training baseline. An FM score close to zero and a comparatively low BWT score suggest minimal forgetting during training and, consequently, strong retention of robustness across drift types. The low negative FWT and high IM score also indicate that exposure to previous drifts overall benefits the learning of new tasks.
ER with a replay buffer size of 5\% still reduces forgetting compared to the FT baseline and outperforms the joint baseline on A-WER, but overall shows greater instability, as evidenced by the large variance, and therefore insufficient replay capacity to stabilize learning across different permutations.

On the other hand, RLR exhibits large forgetting ($-22.34\pm13.32$) and BWT scores, while OGD outperforms RLR across most metrics and achieves the best AI-WER score ($21.19\pm1.18$) overall. OGD enforces orthogonal gradient updates, which we ensure by computing the cosine similarities between the stored task gradients, which are all close to zero, ranging from $10^{-3}$ to $10^{-6}$. This indicates that task gradients lie in distinct subspaces for OGD and suggests that forgetting arises from destructive updates to these gradients rather than from representational drift. However, because we use mean pooling for RLR, which enforces high-level or global similarity, some performance-critical local structures might be missed.

\begin{figure}[t]
    \centering
    \includegraphics[width=\linewidth]{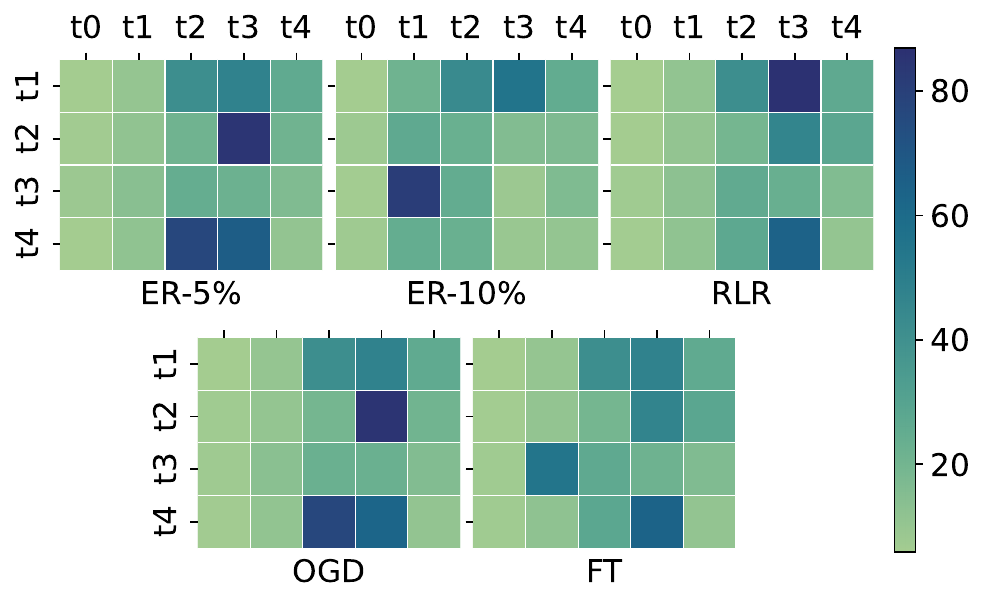}
    \caption{The progression of WER across different methods. Rows correspond to training steps in the curriculum, and columns show the development of per-task performance.}
    \label{fig:cf_matrix_drift}
\end{figure}

If we compare the WER progression during the CL curriculum in Figure~\ref{fig:cf_matrix_drift}, we can see that $t_0$ remains stable and even profits from the sequential training across all methods. For the compound drift $t_4$, we can observe an interesting phenomenon: While the training of $t_4$ mostly hurts existing knowledge, especially for $t_2$ and $t_3$, for $t_1$, the learning of compound drift factors can be beneficial and even restore performance. Also noteworthy is that, for ER-10\%, overall performance recovers.

Overall, we find that robustness across drift types and permutations benefits most from moderately sized memory and replay mechanisms. The high IM scores and large negative FWT scores across all methods, including the FT baseline, indicate that drift types benefit from sequential updates, which is also supported through Figure~\ref{fig:cf_matrix_drift}. However, the large variance in FM and BWT scores across all methods except ER-10\% indicates that the task order must be carefully chosen.

\section{Conclusion}
We introduced MoDiCoL, a Modular Diagnostic Continual Learning dataset for studying ASR robustness to distributional shifts. Built with a factorial design, MoDiCoL evaluates the impact of linguistic content, speaker characteristics, and acoustic environment drift on performance.
We created distinct run configurations with real and synthetic speech, augmented to match defined factor levels, to create a continual learning (CL) curriculum across different drift types. To evaluate adaptation, forgetting, and interference in the model, we applied three CL methods: Experience Replay (ER), Representation-level Regularization (RLR), and Orthogonal Gradient Descent (OGD).

ER with a buffer of 10\% of previous task data stabilized training on new tasks and even outperformed the joint training baseline, indicating that replay is beneficial for preserving robustness across drifts. Using RLR and OGD, we examined whether forgetting was due to interfering parameter updates or representational drift. Since OGD outperformed RLR, which exhibited substantial forgetting, we conclude that interference within gradient subspaces is one factor contributing to performance loss. We found that a sequential introduction of drifts improved the model’s learning plasticity, whereas memory stability was highly dependent on task order and the CL method.

Overall, we showed that CL can not only preserve model robustness to distributional drift but also serve as a diagnostic tool for investigating forgetting mechanisms in pretrained models. We make MoDiCoL, its augmentation pipeline, and the CL curriculum available on Hugging Face. In the future, we plan to expand it with additional runs to include more variations of real-world data, enabling further analysis of model behavior.

\section{Acknowledgments}
The authors gratefully acknowledge funding from Horizon Europe under the MSCA grant agreements No 101072488 (TRAIL), No 101226624 (GREET), and No 101168792 (SWEET).

\section{Generative AI Use Disclosure}
Generative AI tools have been used to polish the language in the manuscript.

\bibliographystyle{IEEEtran}
\bibliography{mybib}

\end{document}